# Planning with Discrete Harmonic Potential Fields

Ahmad A. Masoud
*KFUPM*
*Saudi Arabia*

## 1. The Harmonic Potential Field Planning Approach: A Background

Utility and meaning in the behavior of an agent are highly contingent on the agent's ability to semantically embed its actions in the context of its environment. Such an ability is cloned into an agent using a class of intelligent motion controllers that are called motion planners. Designing a motion planner is not an easy task. A planner is the hub that integrates the internal environment of an agent (e.g. its dynamics and internal representation etc.), its external environment (i.e. the work space in which it is operating, its attributes and the entities populating it), and the operator's imposed task and constraints on behavior as one goal-oriented unit. Success in achieving this goal in a realistic setting seems closely tied to the four conditions stated by Brooks which are: intelligence, emergence, situatedness, and embodiment (Brooks, 1991). While designing planners for agents whose inside occupies a simply-connected region of space can be challenging, the level of difficulty considerably rises when the planner is to be designed for agents whose inside no longer occupies a simply-connected space (i.e. the agent is a distributed entity in space). This situation gives rise to sensitive issues in communication, decision making and environment representation and whether a centralize top-down mode for behavior generation should be adopted or a decentralized, bottom-up approach may better suit the situation at hand.
To the best of this authors' knowledge, the potential field (PF) approach was the first to be used to generate a paradigm for motion guidance (Hull,1932); (Hull, 1938). The paradigm began from the simple idea of an attractor field situated on the target and a repeller field fencing the obstacles. Several decades later, the paradigm surfaced again through the little-known work of Loef and Soni which was carried out in the early 1970s (Loef , 1973); (Loef & Soni, 1975). Not until the mid-1980s did this approach achieve recognition in the path planning literature through the works of Khatib (Khatib , 1985), Krogh (Krogh, 1984), Takegaki and Arimoto (Takegaki & Arimoto, 1981), Pavlov and Voronin (Pavlov & Voronin, 1984) ,Malyshev (Malyshev, 1980), Aksenov et al. (Aksenov et al., 1978), as well as Petrov and Sirota (Petrov, Sirota ,1981); (Petrov & Sirota, 1983). Andrews and Hogan also worked on the idea in the context of force control (Andrews & Hogan, 1983).
Despite its promising start, the attractor-repeller paradigm for configuring a potential field for use in navigation faced several problems. The most serious one is its inability to guarantee convergence to a target point (the local minima problem). However, the problem was quickly solved. One solution uses a configuration that forces the divergence of the PF gradient to be zero almost everywhere in the workspace, hence eliminating the local minima problem. The approach was named the harmonic potential field (HPF) planning approach. A basic setting of this approach is shown in (1) below:

$$\nabla^2 V(X) = 0 \quad X \in \Omega \tag{1}$$

subject to: $V(X_S) = 1$, $V(X_T) = 0$, and $\dfrac{\partial V}{\partial \mathbf{n}} = 0$ at $X = \Gamma$,

where $\Omega$ is the workspace, $\Gamma$ is its boundary, $\mathbf{n}$ is a unit vector normal to $\Gamma$, $X_s$ is the start point, and $X_T$ is the target point.

Although a paradigm to describe motion using HPFs has been available for more than three decades, it was not until 1987 that Sato (Sato, 1987) formally used it as a tool for motion planning (an English version of the work may be found in (Sato, 1993)). The approach was formally introduced to the robotics and intelligent control literature through the independent work of Connolly et al. (Connolly et al., 1990), Prassler (Prassler, 1989) and Tarassenko et al. (Tarassenko & Blake, 1991) who demonstrated the approach using an electric network analogy, Lei (Lei, 1990) and Plumer (Plumer, 1991) who used a neural network setting, and Keymeulen et al. (Decuyper & Keymeulen, 1990); (Keymeulen & Decuyper, 1990) and Akishita et al. (Akishita et al., 1990) who utilized a fluid dynamic metaphor in their development of the approach. Cheng et al. (Cheng & Tanaka, 1991);(Cheng, 1991);(Kanaya et al., 1994) utilized harmonic potential fields for the construction of silicon retina, VLSI wire routing, and robot motion planning. A unity resistive grid was used for computing the potential. In (Dunskaya & Pyatnitskiy 1990) a potential field was suggested whose differential properties are governed by the inhomogeneous Poisson equation for constructing a nonlinear controller for a robotic manipulator taking into consideration obstacles and joint limits.

Harmonic potential fields (HPFs) have proven themselves to be effective tools for inducing in an agent an intelligent, emergent, embodied, context-sensitive and goal-oriented behavior (i.e. a planning action). A planning action generated by an HPF-based planner can operate in an informationally-open and organizationally-closed mode; therefore, enabling an agent to make decisions on-the-fly using on-line sensory data without relying on the help of an external agent. HPF-based planners can also operate in an informationally-closed, organizationally-open mode (Masoud, 2003) ;(Masoud & Masoud, 1998) which makes it possible to utilize existing data about the environment in generating the planning action as well as illicit the help of external agents . A hybrid of the two modes may also be constructed. Such features make it possible to adapt HPFs for planning in a variety of situations. For example in (Masoud & Masoud, 2000) vector-harmonic potential fields were used for planning with robots having second order dynamics. In (Masoud, 2002) the approach was configured to work with a pursuit-evasion planning problem, and in (Masoud & Masoud, 2002) the HPF approach was modified to incorporate joint constraints on regional avoidance and direction. The decentralized, multi-agent, planning case was tackled using the HPF approach in (Masoud, 2007). The HPF approach was also found to facilitate the integration of planners as subsystems in networked controllers containing sensory, communication and control modules with a good chance of yielding a successful behavior in a realistic, physical setting (Gupta et al., 2006).

Although a variety of provably-correct, HPF-based planning techniques exist for the continuous case, to the best of this author's knowledge, there are no attempts to formally use HPFs to synthesize planners that work with discrete spaces described by weighted graphs. Planning in discrete spaces is of considerable significance in managing the complexity in high dimensions (Aarno et al., 2004); (Kazemi et al., 2005). It is also important for dealing with inherently discrete systems such as robustly planning the motion of data packets in a network of routers (Royer & Toh, 1999).

In this work a discrete counterpart to the continuous harmonic potential field approach is suggested. The extension to the discrete case makes use of the strong relation HPF-based planning has to connectionist artificial intelligence (AI). Connectionist AI systems are networks of simple, interconnected processors running in parallel within the confines of the environment in which the planning action is to be synthesized. It is not hard to see that such a paradigm naturally lends itself to planning on weighted graphs where the processors may be seen as the vertices of the graph and the relations among them as its edges. Electrical networks are an effective realization of connectionist AI. Many computational techniques utilizing electrical networks do exist (Blasum et al., 1996);(Duffin, 1971);(Bertsekas, 1996);(Wolaver, 1971) and the strong relation graph theory has to this area (Bollabas, 1979);(Seshu, Reed, 1961) is well-known. This relation is directly utilized for constructing a discrete counterpart to the BVP in (1) used to generate the continuous HPF. The discrete counterpart is established by replacing the Laplace operator with the flow balance operator represented by Krichhoff current law (KCL) (Bobrow, 1981). As for the boundary conditions, they are applied in the same manner as in (1) to the boundary vertices. The discrete counterpart is supported with definitions and propositions that helps in utilizing it for developing motion planners. The utility of the discrete HPF (DHPF) approach is demonstrated in three ways. First, the capability of the DHPF approach to generate new, abstract, planning techniques is demonstrated by constructing a novel, efficient, optimal, discrete planning method called the M* algorithm. Also, its ability to augment the capabilities of existing planners is demonstrated by suggesting a generic solution to the lower bound problem faced by the A* algorithm. The DHPF approach is shown to be useful in solving specific planning problems in communication. It is demonstrated that the discrete HPF paradigm can support routing on-the-fly while the network is still in a transient state. It is shown by simulation that if a path to the target always exist and the switching delays in the routers are negligible, a packet will reach its destination despite the changes in the network which may simultaneously take place while the packet is being routed. An important property of the DHPF paradigm is its ability to utilize a continuous HPF-based planner for solving a discrete problem. For example, the HPF-based planner in (Masoud, 2002) may be adapted for pursuit-evasion on a grid. Here a note is provided on how the continuous, multi-agent, HPF-based planner in (Masoud, 2007) may be adapted for solving a form of the sliding block puzzle (SBP).

## 2. HPF and Connectionism:

Learning is the main tool used by most researchers to adapt the behavior of an agent to structural changes in its environment (Tham & Prager, 1993),(Humphrys, 1995),(Ram et al., 1994). The overwhelming majority of learning techniques are unified in their reliance on experience as the driver of action selection. There are, however, environments which an agent is required to operate in that rule-out experience as the only mechanism for action selection. Learning, or the acquisition of knowledge needed to deal with a situation, may be carried-out via hieratical, symbolic reasoning.  Despite the popularity of the symbolic reasoning AI approach, its fitness to synthesize autonomous and intelligent behavior in such types of environments is being seriously questioned (Brooks, 1990); (Brooks, 1991).

Representing an environment as a group of discrete heterogeneous entities that are glued together via a hierarchical set of relations is a long standing tradition in philosophy and science. There is, however, an opposing, but less popular, camp to the above point of view stressing that representations should be indivisible, and homogeneous. Distributed representations have already found supporters among modern mathematicians, system theorists, and philosophers. Norbert Wiener said "The identity of a body is more like the identity of a flame than that of a stone; it is the identity of a structure, not of a piece of matter" (Wiener, 1950);(Wiener, 1961).  In (Lefebvre,1977) Lefebvre viewes an entity or a process as a wave that glides on a substrate of parts where the relation between the two is that of a system drawn on a system.  In (Campbell , 1994) Campbell argues against the hypothesis that geometrical symbols are used by creatures, to model the environment that they want to navigate. He postulate the existence of a more subtle and distributed representation of the environment inside an agent.  With this in mind, the following guidelines are used for constructing a representation:

1. A representation is a pattern that is imprinted on a substrate of some kind.
2. The substrate is chosen as a set of homogeneous, simple, automata that densely covers the agent's domain of awareness. This domain describes the state of the environment and is referred to as state space.
3. The representation is self-referential. A self-referential representation may be constructed using a dense substrate of automata that depicts the manner in which an agent acts at every point in state space. Self-referential representations are completely at odd with objective representations. They are a product of the stream of philosophy and epistemology (theory of knowledge) (Glasserfeld, 1986), (Lewis, 1929), (Nagel & Brandt, 1965), (Masani, 1994) which stresses that ontological (absolute or objective) reality does not exist, and any knowledge that is acquired by the agent is subjective (self-referential.)
4. In conformity with the view that objective reality is unattainable, a representation is looked upon as merely a belief. Its value to an agent is in how useful it is, not how well it represents its outside reality. Therefore, a pattern that evolves as a result of a self-regulating construction is at all phases of its evolution a legitimate representation.

A machine is a two-port device that consists of an operator port, an environment port, and a construction that would allow a goal set by the operator, defined relative to the environment to be reached. CYBERNETICS (Wiener, 1950); (Wiener, 1961), or as Wiener defined it: "communication and control in the animal and the machine," is based on the  conjectures that a machine can learn, can produce other machines in its own image, and can evolve to a degree where it exceeds the capabilities of its own creator. It is no longer necessary for the operator to generate a detailed and precise plan to convert the goal into a successful motor action. The operator has to only provide a general outline of a plan and the machine will fill in the "gaps"; hence confining the operator's intervention to the high-level functions of the undergoing process. Such functions dictate goals and constrain behavior. The machine is supposed to transform the high-level commands into successful actions. CYBERNETICS unifies the nature of communication and control. It gives actions the soft nature of information. To a cybernist a machine that is interacting with its environment is an agent that is engaging in information exchange with other agents in its environment. In turn, a machine consists of interacting subagents, and is an interactive subagent in a larger machine. A controller which forces an agent to comply with the will of the operator is seen as an encoder that translates the requests of the operator to a language the agent can understand. Therefore, an action is a message, and a message is an information-bearing signal or simply information. Accepting the above paves the way for a qualitative understanding of the ability of a machine to complement the plan of the operator. Let an information theoretic approach (Shannon, 1949); (Gallager, 1968) be used to examine two agents that are interacting or, equivalently, exchanging messages. Assume that the activities of the first agent has $I_x$ equivalence of information, and that the second has $I_y$. Although what is being contributed by the interacting agents is equal to $I_x+I_y$ (self-information), the actual information content of the process is $I_x+I_{xy}+I_y$, where $I_{xy}$ is called mutual information. While the measure of self information is always positive definite ($I_x=-\log(P_x)$, $I_y=-\log(P_y)$), the measure of mutual information ($I_{xy}= \log(P_{x,y}/(P_x.P_y))$) is indefinite ($P_x$ and $P_y$ are the probability of x and y respectively, and $P_{x,y}$ is their joint probability). In an environment where carefully designed modes of interaction are instituted among the constituting agents, the net outcome from the interaction will far exceed the sum of the individual contributions. On the other hand, in non-cooperative environments the total information maybe much less than the self-information (an interaction that paralyzes the members makes $P_{x,y} \equiv 0$, and $I_{xy} \rightarrow -\infty$). It has been shown experimentally and by simulation that sophisticated goal-oriented behavior can emerge from the local interaction of a large number of participants which exhibit a much more simplistic behavior. This has motivated a new look at the synthesis of behavior that is fundamentally different from the top-bottom approach which is a characteristic of classical AI. Artificial Life (AL) (Langton, 1988) approaches behavior as a bottom-up process that is generated from elementary, distributed, local actions of individual organisms interacting in an environment. The manner in which an individual interact with others in its loca1 environment is called the Geno-type. On the other hand, the overall behavior of the group (Phenotype, or P-type) evolves in space and time as a result of the interpretation of the Geno-type in the context of the environment. The process by which the P-type develop under the direction of the G-type is called Morphogenesis (Thorn, 1975).

To alter its state in some environment an agent (from now on is referred to as the operator) needs to construct a machine that would interface its goal to its actions. The machine (or interface) function to convert the goal into a sequence of actions that are imbedded into the environment. These actions are designed to yield a corresponding sequence of states so that the final state is the goal state of the operator. The action sequence is called a plan and it is a member of a field of plans (Action field) that densely covers state space so that regardless of the starting point, a plan always exist to propel the agent to its goal. To construct a machine of the above kind the operator must begin by reproducing itself by densely spreading operator-like micro-agents at every point in state space (Figure-1). The only difference between the operator-agent and an operator-like micro-agent is that the state of the Operator evolves in time and space while the state of the micro-agent is stagnant and immobilized to one *a priori* known point in state space. The second part of machine construction is to induce the proper action structure over the micro-agent group. It is obvious that a hierarchical, holistic, centralized approach for inducing structure over the group entails the existence of a central planning agent/s that is/are not operator-like.

Including such an agent in the machine violates organizational closure, i.e. the restrictions on intelligent machines receiving no influx of external intelligence to help them realize their goals. In other words, the agent must be able to lift itself from its own bootstraps. By restricting the forms of the agents constituting the machine to that of the operator, an AL approach does not require the intervention of any external intelligence to help in the construction of the machine. The AL approach, which is decentralized by definition (i.e. no supervisor is needed), requires a micro-agent to locally constrain its behavior (Genotype, or G-type behavior) using the information derived from the states of the neighboring micro-agents (Figure-1). Unlike centralized approaches where each micro-agent has to exert the "correct" action in order to generate a group structure that unifies the micro-agents in one goal-oriented unit, the AL approach only requires the micro-agents only not to exert the "wrong" action that would prevent the operator from proceeding to its goal. Obviously, not selecting the wrong action is not enough, on its own, for each micro-agent to restrict itself to one and only one admissible action that would constitute a proper building block of the global structure that is required to turn the group into a functional unit. In an AL approach, the additional effort (besides that of the G-type behavior) needed to induce the global structure on the micro-agents is a result of evolution in space and time under the guidance of the environment. This interpretation or guidance is what eventually limits each micro-agent to one and only one action that is also the proper component in a functioning group structure.

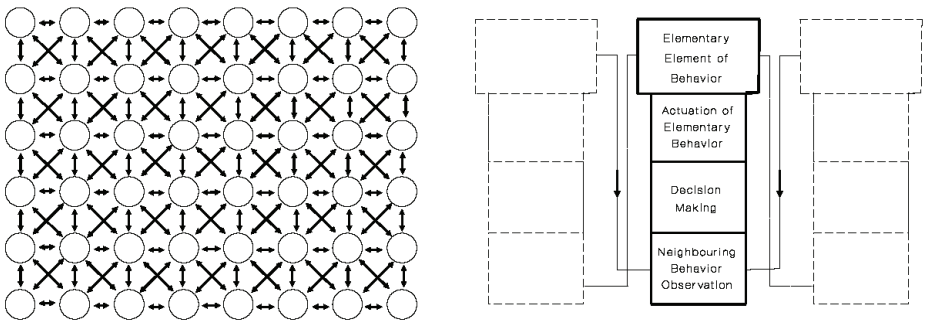

Figure 1. An interacting collective of micro-agents & Layers of functions in a micro-agent

To construct a machine that operates in an AL mode, the operator must have the means to:
1. Reproduce itself at every point in state space.
2. Clone the geno-type behavior in each member of the micro-aqent group.
3. Factor the environment in the behavior generation process.

The HPF approach lends itself to the above guidelines for the construction of an AL-driven, intelligent machine.   The potential field is used to induce a dense collective of virtual agents covering the workspace. The interaction among the agents is generated by enforcing the Laplace equation. The environment is factored into the behavior generation process by enforcing the boundary conditions. More details can be found in (Masoud, 2003);(Masoud & Masoud, 1998).

## 3. The DHPF approach:

This section provides basic definitions and propositions that serve as a good starting point for the understanding and utilization of the DHPF approach.

341

***Definition -1***: Let G be a non-directed graph containing N vertices. Let the cost of moving from vertex i to vertex j be $C_{ij}$ ($C_{ij}=C_{ji}$). Let a potential $V_i$ be defined at each vertex of the graph (i=1,..,N), and $I_{ij}$ be the flow from vertex i to vertex j defined as:

$$I_{ij} = \frac{V_i - V_j}{C_{ij}}, \qquad (2)$$

where $V_i > V_j$. Note that equation-2 is analogous to ohm's law in electric circuits (Bobrow, 1981). Let T and S be the target and start boundary vertices respectively.

A discrete counterpart for the BVP in (1) is obtained if at each vertex of G (excluding the boundary vertices) the balance condition represented by KCL is enforced:

$$\sum_j I_{ij} = 0 \qquad i=1,..,N, \ i \neq T, \ i \neq S$$

and

$$V_S = 1, \ V_T = 0. \qquad (3)$$

***Definition-2***: Let the equivalent cost between any two arbitrarily chosen vertices, i and j, of G ($Ceq_{ij}$) be defined as the potential difference applied to the i-j port of G ($\Delta V$) divided by the flow, I, entering vertex i and leaving vertex j (figure-2)

$$Ceq_{ij} = \frac{\Delta V}{I} \qquad (4)$$

***Proposition-1***: The equivalent cost of a graph, G, that satisfies KCL at all of its nodes (i.e. an electric network) as seen from the i-j port (vertices) is less than or equal to the sum of all the costs along any forward path connecting vertex i to vertex j. Note that if the proposition holds for forward paths, it will also hold for paths with cycles.

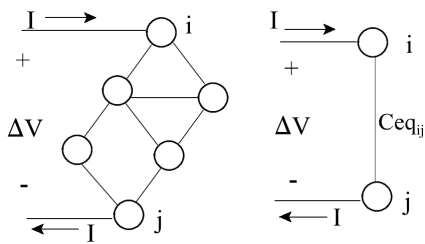

Figure 2. Equivalent cost of a graph as seen from the i-j vertices

***Proof***: Since the graph is required to satisfy KCL, then any internal flow in the edges of the graph is less than or equal to the external flow I (figure-3).

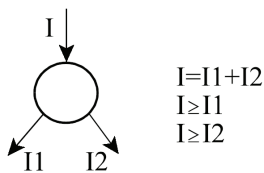

Figure 3. KCL splits input current into smaller or equal components

Consider a forward sequence of vertices connecting vertex i to vertex j: i→i+1→...→i+L-1→j. The potential difference at the k'th edge ($\Delta V_k$) in the selected path is: $\Delta V_k = V_{i+k-1} - V_{i+k}$. The potential difference between i and j may be written as:

$$\Delta V = \Delta V_1 + \Delta V_2 + .... + \Delta V_k + ..... + \Delta V_L \qquad (5)$$

$$= C_1 I_1 + C_2 I_2 + ..... + C_K I_K + .... + C_L I_L ,$$

where for simplicity $C_k$ is used to denote $C_{i+k-1,i+k}$, and $I_k$ is used instead of $I_{i+k-1,i+k}$. Now divide both sides by the flow I:

$$\frac{\Delta V}{I} = Ceq_{i,j} = C_1 \frac{I_1}{I} + ....... + C_k \frac{I_k}{I} + ........ + C_L \frac{I_L}{I} \qquad (6)$$

Since $I_k/I \leq 1$, we have:

$$Ceq_{i,j} \leq C_1 + .... + C_k + .... C_L . \qquad (7)$$

***Proposition-2***: If G satisfies the conditions in equation-3, then the potential defined on the graph (V(G)) will have a unique minimum at T ($V_T$) and a unique maximum at S ($V_S$).

***Proof***: The proof of the above proposition follows directly from KCL and the definition of a flow (equation-2). If at vertex i $V_i$ is a maximum, local or global, with respect to the potential at its neighboring vertices, all the flows along the edges connected to i will be outward (i.e. positive). In this case KCL will fail. Vice versa if $V_i$ is a minimum.

***Proposition-3***: Traversing a positive, outgoing flow from any vertex in G will generate a sequence of vertices (i.e. a path) that terminates at T. Vice versa, traversing a negative, ingoing flow from any vertex in G will generate a sequence of vertices (i.e. a path) that terminates at S.

***Proof***: Assume that a hop is going to be made from vertex i to vertex i+1 based on a selected outgoing, positive flow $I_{i,i+1}$ (Figure-4).

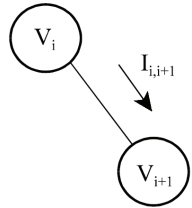

Figure 4. positive flow-guided vertex transition

From the definition of a flow we have:

$$V_i - V_{i+1} = C_{i,i+1} \cdot I_{i,i+1} . \qquad (8)$$

Since we are dealing with positive costs, and the flow that was selected is positive, we have:

$$V_i - V_{i+1} > 0, \quad \text{or} \quad V_i > V_{i+1}. \qquad (9)$$

By guiding vertex transition using positive flows, a continuous decrease in potential is established. This means that the unique minimum, $V_T$, will finally be achieved. In other words, the path will converge to T. The proof of the second part of the preposition can be easily carried out in a manner similar to the first part.

*Proposition-4*: A path linking S to T generated by moving from a vertex to another using a positive flow cannot have repeated vertices (i.e. it contains no loops).
*Proof:* If the flow used to direct vertex transition loops back to a previously encountered vertex, it will keep looping back to that vertex trapping the path into a cycle and preventing it from reaching the target vertex. Since in proposition-3 convergence to T of a positive flow driven path is guaranteed, no cycles can occur and the path cannot have repeated vertices.
*Definition-3*: Since a positive flow path (PFP) beginning at S is guaranteed to terminate at T with no repeated vertices in-between, the combination of all PFPs define a tree with S as the top parent vertex and bottom, offspring vertices equal to T. This tree is called the harmonic flow tree (HFT).
*Proposition-5*: The HFT of a graph contains all the vertices in that graph.
*Proof:* Since all PFPs of a graph start at S and terminate at T, the boundary vertices have to be a part of the HFT. Since at the remaining intermediate vertices the flow balance relation in equation-2 (KCL) is enforced, positive flow must enter each of these vertices; otherwise, the balance relation cannot be enforced. Therefore, every intermediate vertex has to belong to a PFP which makes it also a vertex in the HFT tree.
*Proposition-6*: The HFT of a graph contains the optimal path linking S to T.
*Proof:* If the optimum path of a graph linking S to T is not a branch of its HFT, then some of the hops used to construct that path were driven by negative flows. It is obvious that the optimum path cannot be constructed, even partially, using negative flows. If negative flows are used, convergence cannot be guaranteed. Even if countermeasures are taken to prevent a cycle from persisting, vertices will get repeated and the path will contain loops that can be easily removed to construct a lower cost path. However, a proof that negative flows cannot be used in constructing a least cost path may be obtained using the weighted Jensen inequality:

$$w_1 \cdot f(x_1) + .... + w_L \cdot f(x_L) \geq f(w_1 \cdot x_1 + .... + w_L \cdot x_L) \tag{10}$$

where $f(\ )$ is a convex function and $w_1+...+w_L=1$, $w_i \geq 0$. Equality will hold if and only if $x_1=x_2=..=x_L$.

Assume that the optimal path connecting S to T contains L hops (L+1 vertices). The cost of that path may be written as:

$$C = C_1 + C_2 + .... + C_L = \Delta V_1 \cdot \frac{1}{I_1} + \Delta V_2 \cdot \frac{1}{I_2} .... + \Delta V_L \cdot \frac{1}{I_L}, \tag{11}$$

where $\Delta V_l$'s and $I_l$'s are defined in the proof of proposition-1. Since KCL is enforced we have:

$$\Delta V_1 + .... + \Delta V_L = V_S - V_T = 1 - 0 = 1. \tag{12}$$

Also, notice that $f(I_l)=1/I_l$ is a convex function of $I_l$. Therefore, the weighed Jensen inequality may be applied to the cost function above,

$$\Delta V_1 \cdot f(I_1) + .... + \Delta V_L \cdot f(I_L) \geq f(\Delta V_1 \cdot I_1 + .... + \Delta V_L \cdot I_L). \tag{13}$$

or

$$\sum_{i=1}^{L} C_i \geq \frac{1}{\sum_{i=1}^{L} \Delta V_i \cdot I_i} = \frac{1}{\sum_{i=1}^{L} C_i \cdot I_i^2} . \tag{14}$$

For the case of an electric network $I_k$ and the corresponding $\Delta V_k$ are related using ohm's law. For a positive cost having a negative $\Delta V_k$ implies a negative $I_k$. Therefore, the inequality will still hold for negative values of $\Delta V_k$. This variational inequality may be used to establish a lower bound on the cost of a path connecting the start and target vertices. It is interesting to notice that the lower bound on the cost is equal to the inverse of the power consumed in traversing a path. According to the inequality, a zero difference between the two can only occur if all the flows along the path are the same. In general, the more variation in the value of the flows along the path the more is the deviation from the optimal cost will be. It is obvious that the extreme case of mixing positive and negative flows in constructing the path cannot lead to the optimum solution.

*Proposition-7*: The optimum path (or any PFP for that matter) must contain at most N vertices.

*Proof*: It is obvious that if the path contains more than N vertices, an intermediate vertex in the path is repeated. Based on the previous propositions, this cannot happen. Therefore transition from S to T must be obtained in N hops or less.

## 4. The M* Algorithm:

It is not hard to see that the DHPF belongs to connectionist AI. Even in its raw form, as shown by preposition-3, the approach is capable of tackling planning problems in a provably-correct manner. While connectionist AI and symbolic reasoning AI are considered to be two separate camps in artificial intelligence, there is a trend to hybridize the two in order to generate techniques that combine the attractive properties of each approach. In this section it is demonstrated that the DHPF approach can work in a hybrid mode. The reason for that is: the flow induced by the connectionist system has a structure which can be reasoned about to enhance or add to the basic capabilities of the DHPF approach. This is demonstrated in this section by suggesting the M* algorithm for finding the minimum cost path between S and T on a graph. By interfacing the connectionst layer to a symbolic layer, the quality of the path which the connectionist stage is guaranteed to find is enhanced.

### 4.1 The M*: direct realization:
The followings are the steps for directly realizing the M* procedure:

**01.** Write the KCL equations for each vertex of the graph

$$\sum_j I_{ij} = 0 \quad i = 1,...,N \tag{15}$$

**02.** From the KCL equations derive the vertex potential update equations:

$$V_i = \sum_{\substack{k=1 \\ k \neq i}}^{N} b_{i,k} V_k \tag{16}$$

**03.** Initialize the variables: $\quad V_S=1; \; V_T=0; \; V_i=1/2 \quad i=1,..,N \quad i \neq S, \quad i \neq T$ \hfill (17)

**04.** Loop till convergence is achieved performing the operations:

$$V_i = \sum_{\substack{k=1 \\ k \neq i}}^{N} b_{i,k} V_k \quad i=1,..,N, i \neq S, \ i \neq T \quad (18)$$

**05.** Compute the flows
**06.** Using the flows construct the HFT of the graph
**07.** Starting from the last parent nodes, for each node retain the branch with lowest cost and delete the others
**08.** Move to the parent nodes one level up and repeat step 7.
**09.** Repeat step 8 till the top parent node S is reached
**10.** The remaining branch connected to S is the optimal path linking S to T.

### 4.2 An Example

Consider the weighted graph shown in figure-5. It is required that a minimum cost path be found from the start vertex S=1 to the target vertex T=5. The transition costs are: $C_{16}=1$, $C_{14}=3$, $C_{23}=4$, $C_{34}=7$, $C_{26}=1$, $C_{37}=5$, $C_{35}=2$, $C_{47}=6$, $C_{67}=9$, $C_{57}=5$. The KCL equations are:

$$\text{Vertex-1: } \frac{V_1-V_4}{C_{14}} + \frac{V_1-V_6}{C_{16}} = 0 \quad \text{Vertex-2: } \frac{V_1-V_4}{C_{14}} + \frac{V_1-V_6}{C_{16}} = 0$$

$$\text{Vertex-3: } \frac{V_3-V_2}{C_{23}} + \frac{V_3-V_4}{C_{34}} + \frac{V_3-V_5}{C_{35}} + \frac{V_3-V_7}{C_{37}} = 0 \quad \text{Vertex-4: } \frac{V_4-V_1}{C_{14}} + \frac{V_4-V_3}{C_{34}} + \frac{V_4-V_7}{C_{47}} = 0$$

$$\text{Vertex-5: } \frac{V_5-V_3}{C_{35}} + \frac{V_5-V_7}{C_{57}} = 0 \quad \text{Vertex-6: } \frac{V_6-V_1}{C_{16}} + \frac{V_6-V_2}{C_{26}} + \frac{V_6-V_7}{C_{67}} = 0 \quad (19)$$

$$\text{Vertex-7: } \frac{V_7-V_3}{C_{37}} + \frac{V_7-V_4}{C_{47}} + \frac{V_7-V_5}{C_{57}} + \frac{V_7-V_6}{C_{67}} = 0.$$

The update equations may be derived as:

$$V_1 = \frac{1}{K_1}[\frac{V_4}{C_{14}} + \frac{V_6}{C_{16}}] = b_{1,4}V_4 + b_{1,6}V_6 \ , \quad V_2 = \frac{1}{K_2}[\frac{V_3}{C_{23}} + \frac{V_6}{C_{26}}] = b_{2,3}V_3 + b_{2,6}V_6$$

$$V_3 = \frac{1}{K_3}[\frac{V_2}{C_{23}} + \frac{V_4}{C_{34}} + \frac{V_5}{C_{35}} + \frac{V_7}{C_{37}}] = b_{3,2}V_2 + b_{3,4}V_4 + b_{3,5}V_5 + b_{3,7}V_7$$

$$V_4 = \frac{1}{K_4}[\frac{V_1}{C_{14}} + \frac{V_3}{C_{34}} + \frac{V_7}{C_{47}}] = b_{4,1}V_1 + b_{4,3}V_3 + b_{4,7}V_7 \ , \quad V_5 = \frac{1}{K_5}[\frac{V_3}{C_{35}} + \frac{V_7}{C_{57}}] = b_{5,3}V_3 + b_{5,7}V_7 \quad (20)$$

$$V_6 = \frac{1}{K_6}[\frac{V_1}{C_{1,6}} + \frac{V_2}{C_{2,6}} + \frac{V_7}{C_{6,7}}] = b_{6,1}V_1 + b_{6,2}V_2 + b_{6,7}V_7$$

$$V_7 = \frac{1}{K_7}[\frac{V_3}{C_{37}} + \frac{V_4}{C_{47}} + \frac{V_5}{C_{57}} + \frac{V_6}{C_{67}}] = b_{7,3}V_3 + b_{7,4}V_4 + b_{7,5}V_5 + b_{7,6}V_6$$

where

$$\frac{1}{K_1}=[\frac{1}{C_{14}}+\frac{1}{C_{16}}] \qquad \frac{1}{K_2}=[\frac{1}{C_{23}}+\frac{1}{C_{26}}] \qquad \frac{1}{K_3}=[\frac{1}{C_{23}}+\frac{1}{C_{34}}+\frac{1}{C_{35}}+\frac{1}{C_{37}}]$$

$$\frac{1}{K_4}=[\frac{1}{C_{14}}+\frac{1}{C_{34}}+\frac{1}{C_{47}}] \qquad \frac{1}{K_5}=[\frac{1}{C_{35}}+\frac{1}{C_{57}}] \qquad \frac{1}{K_6}=[\frac{V_1}{C_{16}}+\frac{V_2}{C_{26}}+\frac{V_7}{C_{67}}]$$

$$\frac{1}{K_7}=[\frac{1}{C_{37}}+\frac{1}{C_{47}}+\frac{1}{C_{57}}+\frac{1}{C_{67}}]$$

Setting $V_1=1$, $V_5=0$ and applying the procedure described above we obtain the vertices potential: $V_1=1$, $V_2=0.74673$, $V_3=0.33753$, $V_4=0.70006$, $V_5=0$, $V_6=0.84902$, $V_7=0.41093$. The flows may be computed as: $I_{14}=0.09998$, $I_{16}=0.15098$, $I_{23}=0.1023$, $I_{47}=0.048189$, $I_{43}=0.05179$, $I_{35}=0.16877$, $I_{62}=0.1023$, $I_{67}=0.048677$, $I_{73}=0.014679$, $I_{75}=0.082186$. The graph, flows and corresponding HFT are in figure-5.

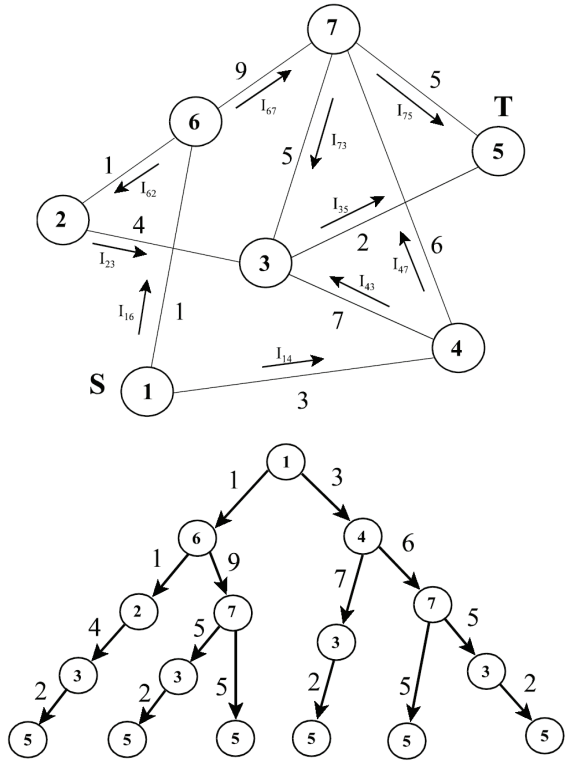

Figure 5. The graph and corresponding flows and HFT

Now start reducing the HFT,

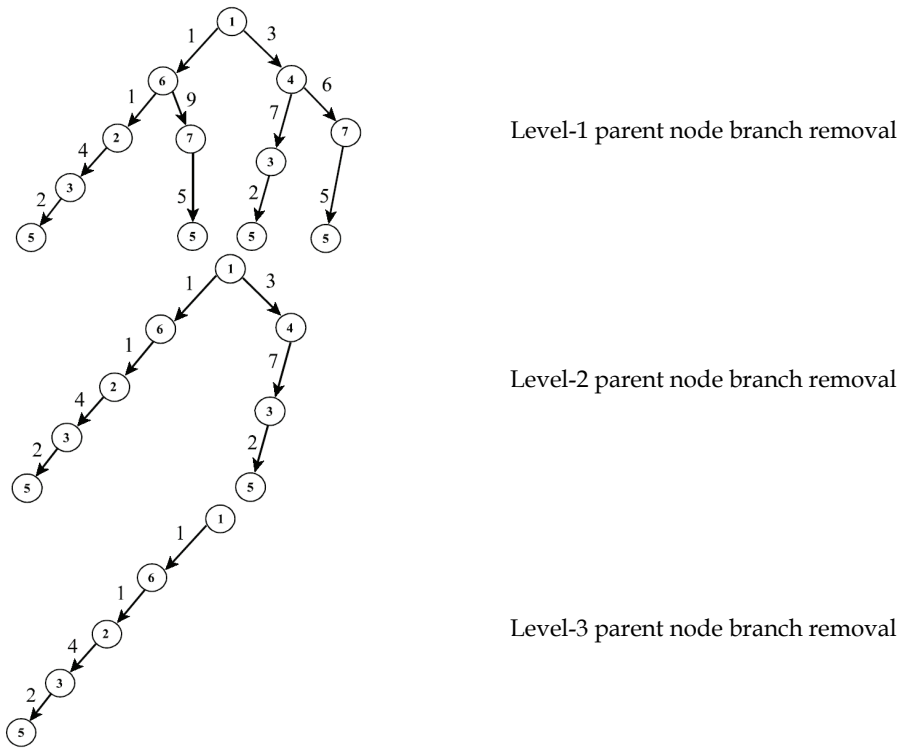

Level-1 parent node branch removal

Level-2 parent node branch removal

Level-3 parent node branch removal

As can be seen from figure-6 the optimum path: 5→ 3→2→6→1 with a cost 8 was obtained after only two levels of branch removal. The HFT need not be computed explicitly in order to carry out the branch removal procedure. The M* algorithm may be implemented indirectly using a procedure that successively relaxes the graph till only the optimal path is left.

**4.2 The M* indirect realization:**
In this subsection a successive relaxation procedure for implementing M* is suggested. The rapid growth of an HFT with the size of a graph makes it impractical to apply the algorithm on the tree directly. Here a procedure that allows us to operate on the HFT indirectly by successively relaxing the graph is suggested. In order to apply the procedure the following terms need to be defined (figure-7): positive flow index (PFI) of a vertex: number of edges connected to the vertex with outward positive flows. Negative flow index (NFI) of a vertex: is the number of edges connected to the vertex with inward negative flows.

| Vertex | {} | {} | {} | {} |
|---|---|---|---|---|
| PFI | 0 | 1 | 2 | 3 |
| NFI | 3 | 2 | 1 | 0 |

Figure 7. PFI and NFI of a vertex

The procedure is:

**0-** compute the PFI and NFI for each vertex of the graph,

**1-** starting from the target vertex and using the negative flows along with the NFIs and PFIs of the vertices, detect the junction vertices and label them based on their levels,

**2-** at the encountered junction vertex clear NB buffers where NB=PFI of the junction vertex,

**3-** starting from the junction vertex, trace forward all paths to the target vertex traversing vertices with positive flows and PFIs=1,

**4-** excluding the lowest cost path, delete all the edges in the graph connecting the junction vertex to the first, subsequent vertices in the remaining paths,

**5-** decrement the PFI of the junction vertex by the number of edges removed from the graph,

**6-** decrement the NFIs of the first subsequent vertices from step 4 by 1,

**7-** if the NFI of any vertex in the graph from step 4 is equal zero and the vertex is not a start vertex, delete the edge in the graph connecting that vertex to the subsequent vertex and reduce the NFI of the subsequent vertex by 1,

**8-** repeat 7 till all the reaming vertices in the paths with PFIs=1 have NFIs >0,

**9-** go to 1 and repeat till there is only one branch left in the graph with two terminal vertices having NFI=0, PFI=1 and NFI=1, PFI=0. This is the optimum path connecting the start vertex to the target vertex.

In figure-8 the procedure is applied, in a step by step manner, to gradually reduce the graph in figure-5 to the optimum path connecting vertex 1 to 5.

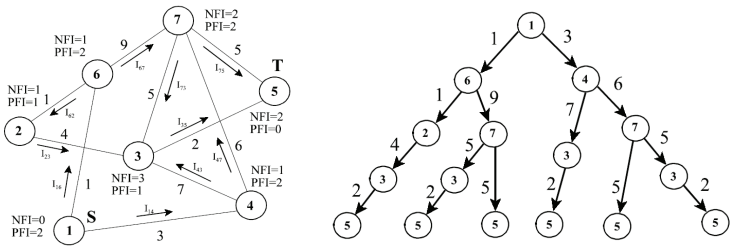

initially traced path: 5→7
constructed paths:  7→5      cost=5
                    7→3→5    cost=5+2=7 (eliminate edge 7→3 in graph)

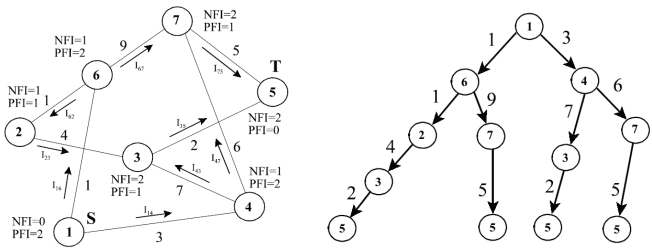

initially traced path: 5→3→2→6
constructed paths:  6→7→5    cost=9+5=14 (eliminate edge 6→7 in graph)
                    6→2→3→5  cost=1+4+2=7

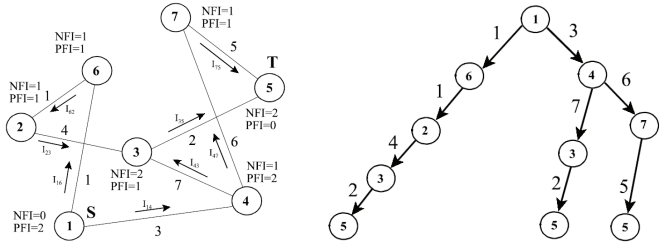

initially traced path: 5→7→4
constructed paths:  4→7→5    cost=6+5=11 (eliminate edge 4→7 in graph)
                    4→3→5    cost=7+2=9

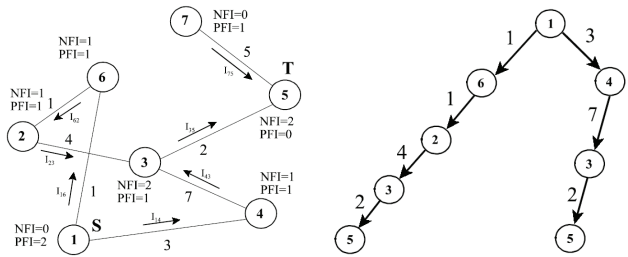

Vertex 7 NFI=0, not a start vertex (eliminate edge 7→5 in graph)

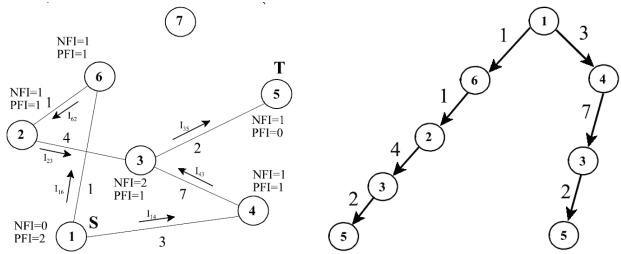

initially traced path: 5→3→2→6→1
constructed paths:  1→4→3→5   cost= 3+7+2=12 (eliminate edge 1→4 in graph)
                    1→6→2→3→5  cost=1+1+4+2=8

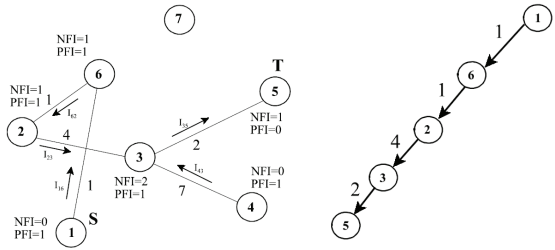

Vertex 4 has an NFI=0 and is not a start vertex (eliminate edge 4 → 3)

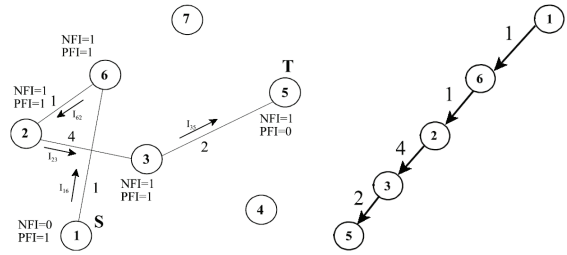

All intermediate vertices have an NFI=PFI=1 → Algorithm terminates.
Figure 8. M*-based successive relaxation.

The optimum path is: 1→6→2→3→5 having a cost of 8.

## 5. A Lower bound for A*

For the A* algorithm to work, a lower bound on the cost from each vertex of the graph to the target vertex has to be supplied. For spatial planning problems the Euclidian distance between the vertices provides such a bound. However, for the general case finding a lower bound may be a source of difficulties that prevents the use of the A* algorithm. In the followings it is shown that the concept of equivalent cost (resistance) from the resistive network paradigm can effectively solve this problem.

Consider the simple graph in figure-9 where S=1 and T=4. To apply the A* algorithm, the path at node 1 should be expanded towards 2 and 3. In order to sort the paths so that the next path expansion can be determined, lower estimates on the cost of moving from 2 to 4 and 3 to 4 are needed. Expansion of the path towards 2 may be achieved by simply removing all the edges of the graph that are attached to 1 leaving only the edge connected to vertex 2 (figure-9). The flows are then computed for the remaining part of the network. The equivalent cost from 2 to 4 ($Ceq_{24}$) may be computed as:

$$Ceq_{24} = \frac{1}{I_{12}} - C_{12} \tag{21}$$

Since in proposition-1 it is proven that the equivalent cost between two vertices in a graph is less than or equal to the least cost path connecting these vertices, the equivalent cost may be used as the lower bound estimate required by the A* algorithm. The minimum cost bounds needed for the remaining path expansions may be obtained in a similar manner.

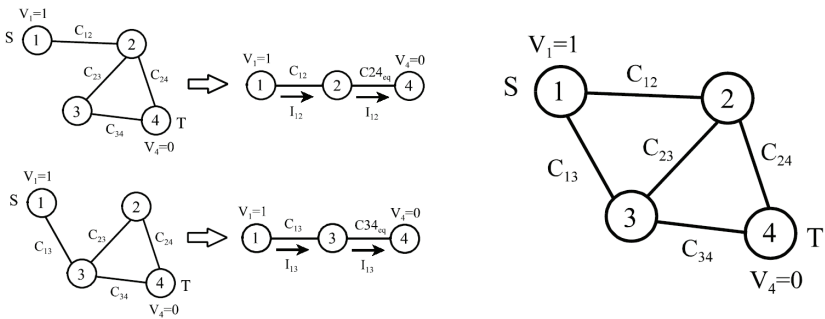

Figure 9. Path expansion, the A*

## 5.1 An example:

The same example in the previous section is repeated using the A* algorithm and the equivalent cost concept. The successive path expansions are shown in figure-10. The optimum path is: 1→6→2→3→5 having a cost of 8.

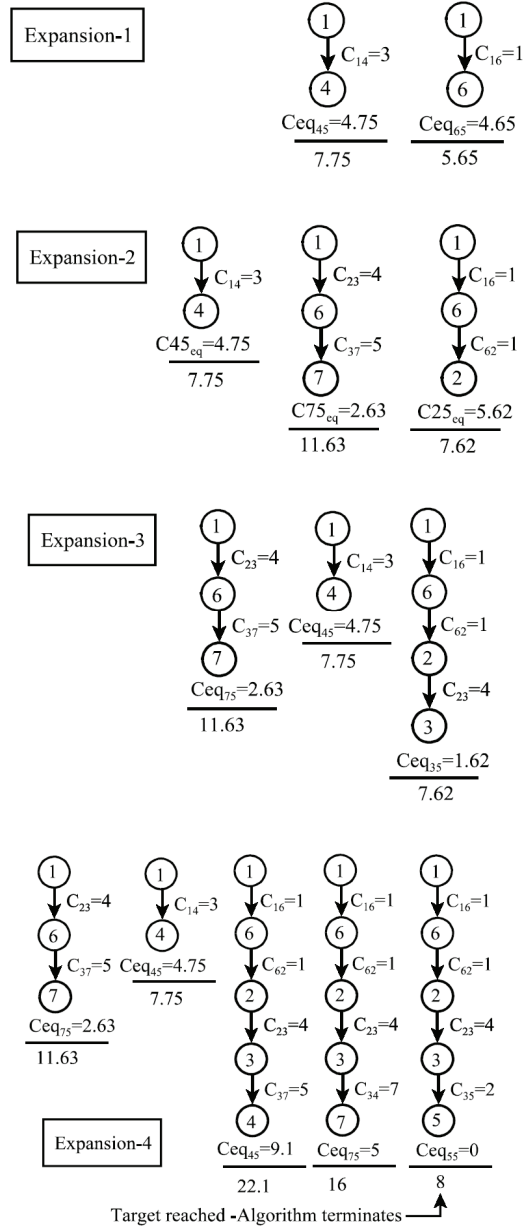

Figure 10. A* applied to the graph in figure-4

## 6. Routing on-the-fly

In the previous sections optimal algorithms for planning motion on a weighted graph utilizing the flow in a resistive grid are suggested. In order to apply these algorithms the graph must have a fixed structure known to the central unit that is processing the data and generating the path. While the above setting applies in many practical situations there are cases where such a scenario cannot be applied, e.g. ad-hoc networks. Also, reliability and cost issues may make it undesirable to have the whole process hinge on the success of a single, central agent. The alternative is to execute the routing process in an asynchronous, decentralized, self-organizing manner. In this case each vertex of the graph is assumed to be a router with limited sensing, processing and decision making capabilities where the immediate domains of awareness and action of a router are limited to a subset of the network with the remaining part being transparent to the router concerned. In other words, the router should sense locally, reason locally, and act locally yet produce global results (figure-11).

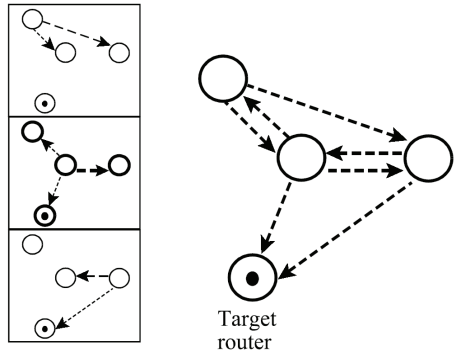

Figure 11. decentralized routing

In a centralized mode, the routers keep exchanging states till convergence is achieved. The potential is then communicated to a central agent which in a single shot lays a path to the target (figure-12). In a decentralized mode, communication of states between routers need not necessarily be sustained till a steady state is reached. Instead, during communication among the routers, whenever possible, the router with the packet attempt to pass it to a neighboring router using a simple, local, potential-based procedure that can be easily implemented on-board a router. As can be seen, under ideal situation, in a discrete HPF paradigm, the decentralized mode reduces to the centralized one.

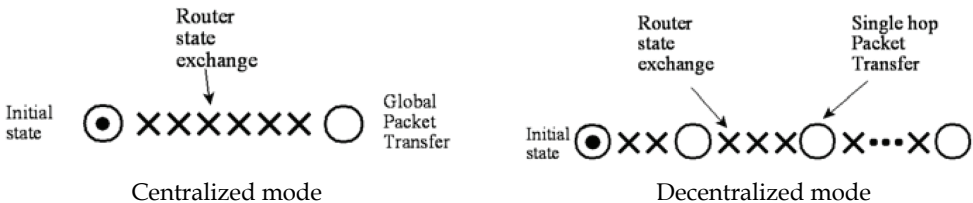

Centralized mode                              Decentralized mode
Figure 12. centralized and decentralized modes in a discrete HPF paradigm

The following is one of the decentralized procedures that may be derived from this paradigm:
**0-** fix the potential at the target vertex to zero,
**1-** each router establishes connectivity with selected neighboring routers and assigns appropriate costs,
**2-** fix the potential at the router that currently hold the data packet to 1,
**3-** excluding the routers with the packet and the target router, each router should update its potential using equation (18),
**4-** forward the packet from the current router to the associated router with highest positive flow,
**5-** if the router is not the target router go to 1,
**6-** target router is reached.

The procedure is simulated for the graph in figure-5. The potential field was initially set using a random number generator that is uniformly distributed between (0, 1). The output of the process is the path: $1 \to 6 \to 2 \to 3 \to 5$ having the cost 8.

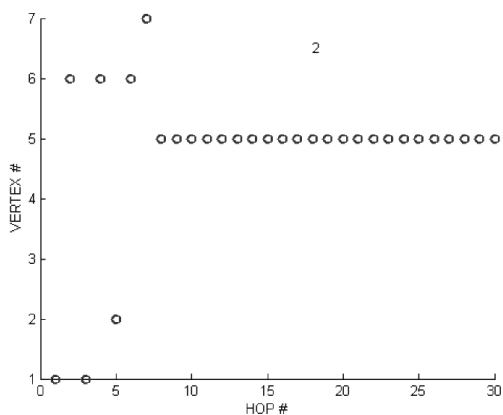

Figure 13. vertex number vs hop number under random router malfunction

To test the robustness of the procedure the example is repeated while inducing, at each hop, a malfunction in a randomly selected router (excluding the target router and the one currently holding the packet). In the following the vertex number as a function of the hop number is shown for one of the trials (figure-13). As can be seen the packet finally converges to the target vertex. Convergence was observed for all the trials that were carried out. The number of hops needed for the packet to reach the target vertex as a function of the trial number and the corresponding histogram are shown in figure-14.

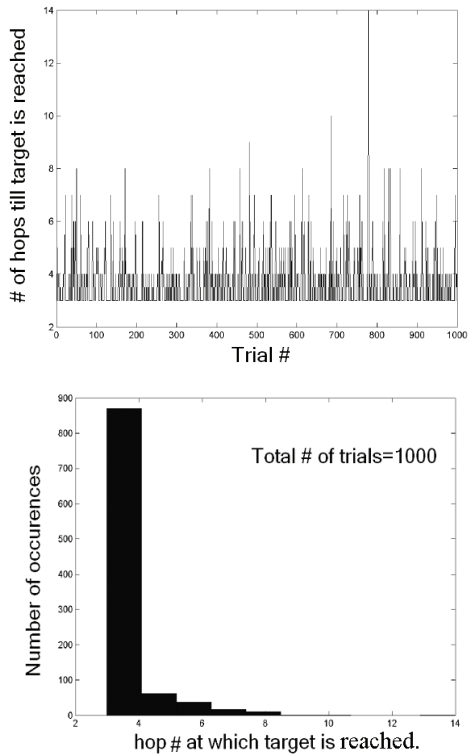

Figure 14. convergence hop number versus trial number and the corresponding histogram

## 7. A Note on Solving the Sliding Block Puzzle:

Designing a planer for the sliding block puzzle (Hordern, 1986) is a challenging task. It is proven that, if symbolic reasoning AI is used, the problem is PSPACE-complete. The difficulty of this problem made it an excellent choice for testing heuristic AI algorithms in order to plan the tiles' motion so that a certain target arrangement is achieved.
With little modification the continuous, multi-agent, decentralized, HPF-based planner suggested in (Masoud, 2007) may be used to tackle specific forms of the SBP problem. This HPF-based planner guarantees that convex shaped blocks can be placed in any configuration within an arbitrarily- shaped confine provided that the narrowest passage within the confine is wide enough to allow the two largest objects to pass at the same time. In other words, if the SBP has two tiles missing, it is completely solvable. The usual form of the SBP contains empty space for one block only. For this case the multi-agent planner cannot guarantee that motion of the tiles can be planned so that a final configuration, even if it belongs to the resolvable set, is attained.
The multi-agent planner uses two types of fields for controlling the motion of each tile of the SBP: a global field, called the purpose field (PRF), constructed using harmonic potential for each agent individually assuming that no other agents is sharing its workspace. This field function to drive the agent to the specified target. The second component is a local field fencing the agents. This component is called the conflict resolving field (CRF). Figure-15 shows the, evolutionary self-organizing nature of the multi-agent, HPF-based planner while attempting to lay trajectories for two sets of robots moving along opposite direction along a tight road.

The method in (Masoud, 2007) may be applied with little modification to the SBP problem. The only modification is to replace the rectangular space in which the tiles of the SBP slide with a rectangular grid (figure-16). The PRF of the individual tile may be easily computed using equation (3). The CRF components remain unchanged.

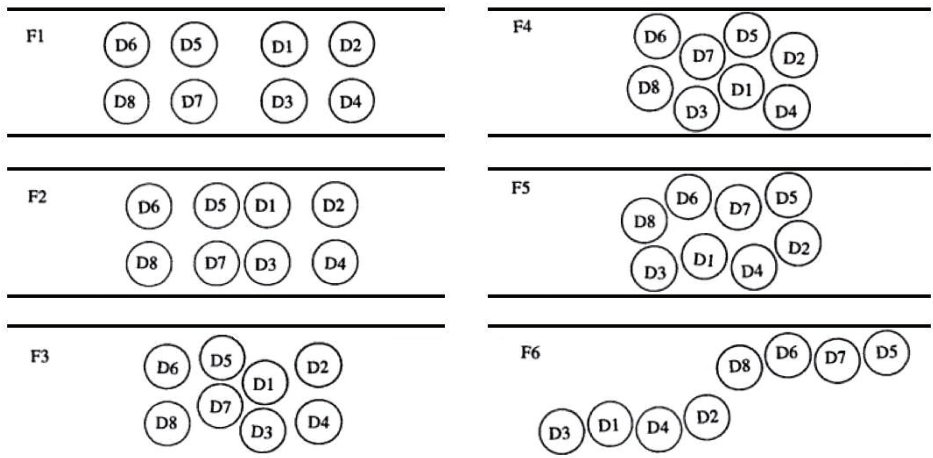

Figure 15. Problem solving through self-organization

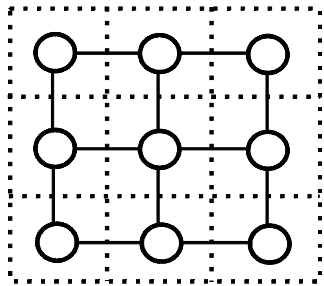

Figure 16. Restricting a PRF to a rectangular grid in a SBP solver

## 8. Conclusions

This work is an attempt to show that in addition to the attractive features the HPF approach already have, it is fully capable of tackling planning problems in discrete spaces. The ability to extend the approach to such type of problems is not a mere coincidence; it is rather the product of a deeply rooted relation the HPF approach has to both AI and mathematics. This relation is what allows this approach to provide a framework able to accommodate a large variety of planning situations, both applied and abstract, as well as provide the tools needed to analyze them and understand their behavior in both qualitative and quantitative manners. Besides the abstract side, the HPF approach possess a physical intuitive side that serves as a good aid for configuring the approach to suit a desired planning situation as well as help in understanding the properties of the suggested method.

The ability of the HPF approach to work in a hybrid, symbolic-connectionist, AI mode adds more value to this method. This value is vivid in the case of data network where communication can still be maintained through local interaction even if the central entity planning traffic is no longer functioning. The fact that in this work the development for the DHPF approach is only provided for non-directed graphs does by no means reflect any shortcoming on part of the approach to deal with directed graphs. The ability of the HPF approach to enforce directional constraints for both the continuous and discrete cases has already been demonstrated in (Masoud & Masoud, 2002). The DHPF approach is even capable of making the connectivity between the vertices conditional on external, user-specified events while providing a provably-correct planning action.

This author is a firm believer in the point of view stating that a the discrete and continuous components in a hybrid system are inseparable (Wiener, 1961); (Wiener, 1950); (Lefebvre; 1977). Rather, the discrete system should appear as a pattern induced on a continuous substrate instead of a distinct discrete system module interfaced to a continuous one. This work and the associated mathematical development should be looked at only as a proof-of-principle of the ability of the HPF approach to tackle, in a provably-correct way, planning in purely discrete domains. This author views this proof of principle as an important argument against the misconception that the HPF approach is incapable of doing so.